\documentclass[sigconf]{acmart}

\usepackage{amsmath}
\usepackage{amsthm}
\usepackage{subfig}

\usepackage{framed}
\usepackage{graphicx}
\usepackage{color}
\usepackage{multirow}
\usepackage[noend]{algorithmic}
\usepackage{algorithm}

\AtBeginDocument{%
  }

\acmConference[UDM-KDD'23
]{2nd KDD Workshop on Uncertainty Reasoning and Quantification in Decision Making}{August 07,
  2023}{Long Beach, CA}

\begin{document}

\title{Uncertainty Quantification of Deep Learning for Spatiotemporal Data: Challenges and Opportunities}

\author{Wenchong He}
\email{whe2@ufl.edu}
\affiliation{%
  \institution{University of Florida}
  \city{Gainesville}
  \state{FL}
  \country{USA}
}

\author{Zhe Jiang}
\email{zhe.jiang@ufl.edu}
\affiliation{%
  \institution{University of Florida}
  \city{Gainesville}
  \state{FL}
  \country{USA}}

\renewcommand{\shortauthors}{He et al.}

\begin{abstract}
With the advancement of GPS, remote sensing, and computational simulations, large amounts of geospatial and spatiotemporal data are being collected at an increasing speed. Such emerging spatiotemporal big data assets, together with the recent progress of deep learning technologies, provide unique opportunities to transform society. However, it is widely recognized that deep learning sometimes makes unexpected and incorrect predictions with unwarranted confidence, causing severe consequences in high-stake decision-making applications  (e.g., disaster management, medical diagnosis, autonomous driving). Uncertainty quantification (UQ) aims to estimate a deep learning model's confidence. This paper provides a brief overview of  UQ of deep learning for spatiotemporal data, including its unique challenges and existing methods. We particularly focus on the importance of uncertainty sources. We  identify several future research directions for spatiotemporal data.
\end{abstract}

\maketitle

\section{Introduction}
With the advancement of GPS, remote sensing, and computational simulations, large amounts of geospatial and spatiotemporal data are being collected at an increasing speed~\cite{shekhar2015spatiotemporal}. Such emerging spatiotemporal big data assets, together with the recent progress of deep learning technologies, provide unique opportunities to transform society in broad applications. 
For example, deep learning is widely used to process spatiotemporal data from radar or lidar sensors and video cameras to monitor road conditions, detect pedestrians, and navigate through traffic \cite{zhao2019t}. In disaster management, deep learning systems have been developed to analyze satellite or drone imagery to enhance situational awareness during deadly hurricane flood disasters~\cite{jiang2019spatial}. Although deep learning is known for higher prediction accuracy compared with many traditional machine learning techniques, it is widely recognized that they can sometimes make unexpected and incorrect predictions with unwarranted confidence, particularly in complex real-world environments \cite{reichstein2019deep}. This can have serious consequences in high-stakes applications like autonomous driving \cite{choi2019gaussian}, medical diagnosis \cite{begoli2019need}, and disaster response \cite{alam2017image4act}. Therefore, uncertainty quantification is essential for a deep learning model to be aware of its limitations and avoid overconfident predictions. 

{\bf Applications}: One important application is disaster response. Deep learning has been used to predict the track of hurricanes or estimate flooded areas, the results of which directly impact decision-makers in planning evacuation and rescue efforts. Thus, in hurricane tracking, scientists often provide not only the most likely point of landfall but also provide a “cone of uncertainty” across other likely points of impact and future trajectories of the storm. Similarly, in autonomous driving, complex environmental factors like extreme weather or the ambiguous appearance of nearby vehicles can fool a deep learning model to ignore an obstacle and cause traffic crashes. In the medical domain, deep learning has been widely used for medical image analysis, clinical diagnosis, and treatment planning. Overconfident predictions  can not only cause unnecessary medical expenses but also endanger patient life \cite{loftus2022uncertainty}.

{\bf Challenges:} Uncertainty quantification of deep learning for spatiotemporal data poses unique challenges due to their special data characteristics.  First, spatiotemporal data violate the common assumption that samples follow an identical and independent distribution. 
Instead, implicit dependency structures exist in continuous space and time (e.g., spatial and temporal autocorrelation, and temporal dynamics)~ \cite{shekhar2015spatiotemporal,jiang2018survey,he2022explainer}. Thus, the uncertainty quantification process should be aware of such a dependency structure. Second, spatiotemporal data have various spatial, temporal, and spectral resolutions and diverse sources of noise and errors (e.g., sensor noise, obstacles, and atmospheric effects in remote sensing signals~\cite{licata2022uncertainty}, GPS errors). 
Analyzing such data often requires the co-registration of different layers (e.g., points, lines, polygons, geo-rasters) into the same spatial reference system. The process is subject to registration uncertainty due to GPS errors or annotation mistakes in map generation~ \cite{jiang2022weakly,he2022quantifying}.
Third, spatiotemporal data are heterogeneous (non-stationary), i.e., the data distribution often varies across different regions or time periods \cite{jiang2019spatial}. Thus, a deep learning model trained in one region (or time) may not generalize well to another region (or time). Spatiotemporal non-stationary requires characterizing uncertainty due to out-of-distribution samples~\cite{shekhar2015spatiotemporal}. This issue is particularly important when spatial observation samples are sparsely distributed, causing uncertainty when inferring the observations at other locations in continuous space \cite{hengl2017soilgrids250m}. 
Moreover, modeling such uncertainty needs to consider sample density both in  the non-spatial feature space and in the geographic space. 
Fourth, the large volume of spatiotemporal data requires efficient computation for prediction and uncertainty quantification, especially considering the need of capturing long-range and complex dependencies. 
Finally, spatiotemporal phenomena are often governed by physical laws (e.g., water flow, temperature diffusion), and it is crucial to consider the inherent physics knowledge of the system when quantifying prediction uncertainty for spatiotemporal systems.

This paper does not aim to provide a thorough review of uncertainty quantification methods for deep learning, as this can be found in several existing surveys \cite{gawlikowski2021survey, he2023survey}. Instead, the paper provides a brief overview of uncertainty quantification methods for spatiotemporal data. We first categorize the sources of uncertainty into two types: data uncertainty and model uncertainty. We delve into these sources and explore how they can be effectively represented in neural network models. We briefly summarize existing methodologies for quantifying uncertainty in deep neural networks, taking into account the perspective of uncertainty sources. We analyze the advantages and disadvantages of these methodologies, shedding light on their applicability and limitations in the context of spatiotemporal data. Finally, we identify several future research directions related to uncertainty quantification for spatiotemporal data.

\section{Types of uncertainty source}

In this section  categorize the sources of uncertainty in GeoAI into two types: data uncertainty and model uncertainty. We discuss the potential sources of each type and its representation.


 \subsection{Data uncertainty}
\subsubsection{Source of data uncertainty}

\begin{figure*}
    \centering
   \subfloat[Classes distribution with clean separable boundary]{ \includegraphics[height=0.95in]{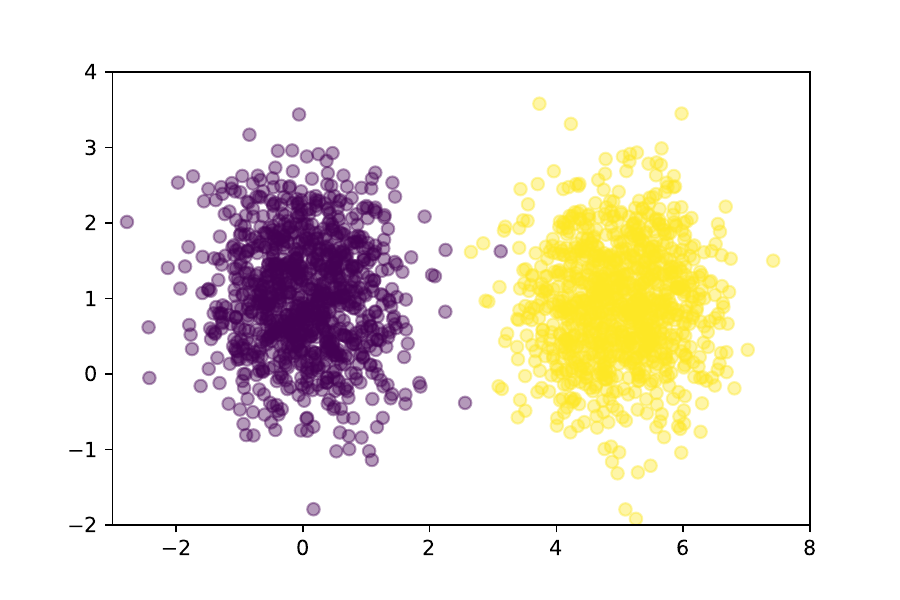}}
   \subfloat[Entropy of samples in (a)]{\includegraphics[height=0.95in]{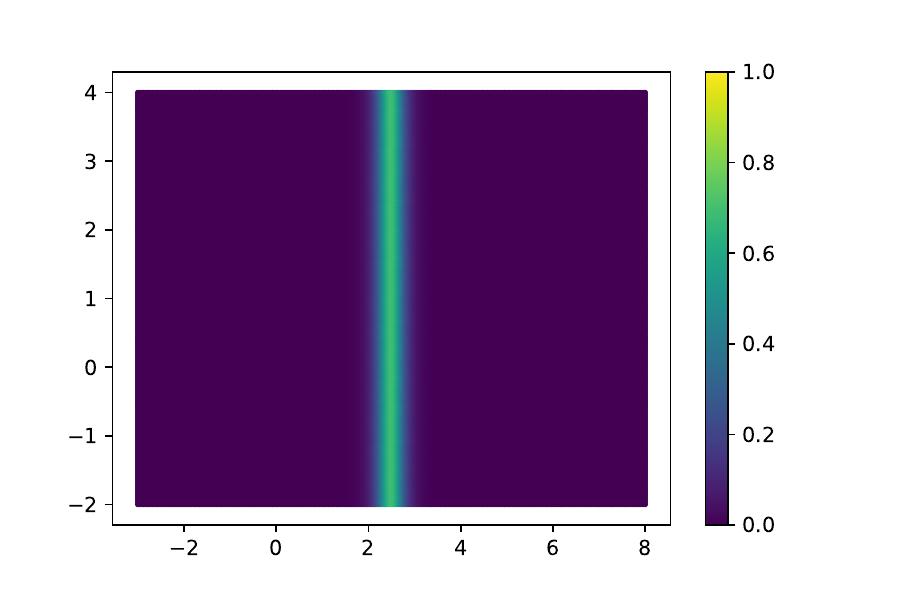}} 
   \subfloat[Classes distribution with ambiguous boundary]{\includegraphics[height=0.95in]{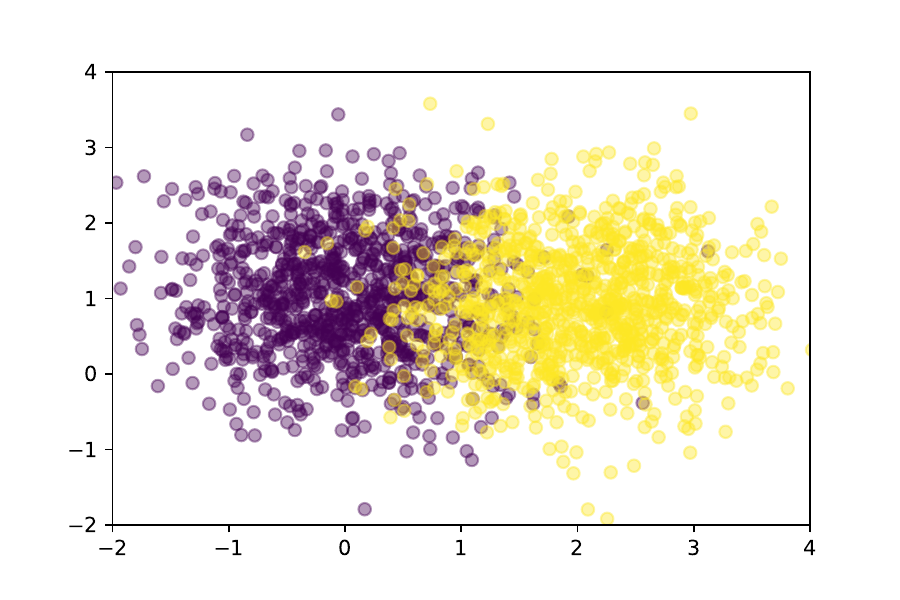}}
   \subfloat[Entropy of samples in (c)]{\includegraphics[height=0.95in]{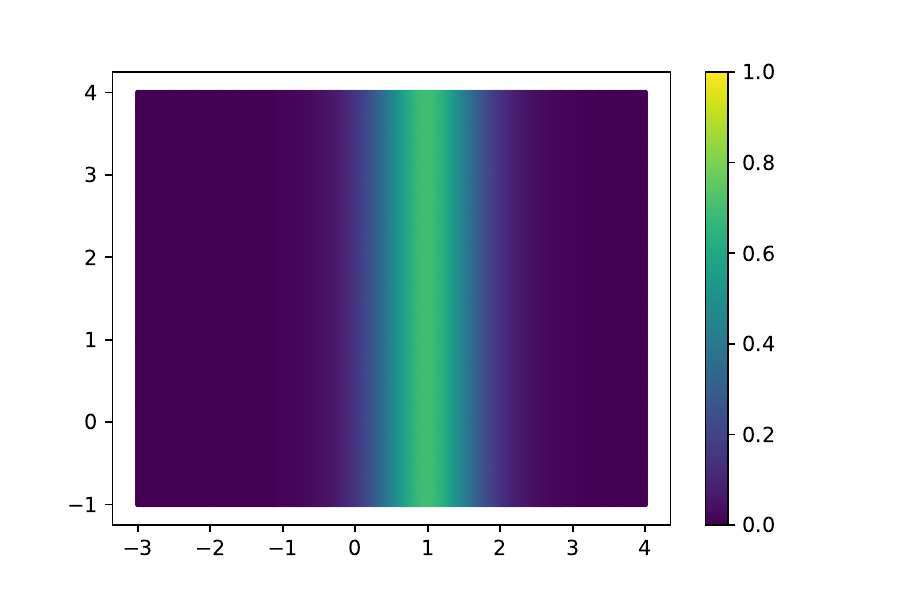}}
    \caption{Data uncertainty visualization examples (Different colors represent samples in different classes)}
    \label{fig:entropy_uq}
\end{figure*}

Data uncertainty, also known as aleatoric uncertainty, arises from the inherent randomness, noise, or overlapping feature distribution within the data, which cannot be eliminated even with additional training data \cite{yarin2016uncertainty}. Randomness or noise in data can stem from various factors during data acquisition. Uncertainty in the data acquisition process can be attributed to instrument errors, inadequate data sampling frequency, and transmission errors \cite{hariri2019uncertainty}. Complex environmental conditions, such as poor weather, can hinder the proper functioning of data acquisition devices. Additionally, uncertainty can be influenced by inappropriate sampling methods, data storage, data representation techniques, and interpolation methods during data processing \cite{yarin2016uncertainty}.

In the case of spatiotemporal data collected from space and airborne platforms like CubeSats and UAVs, data uncertainty can arise not only from sensor errors in data acquisition devices but also from the discretization of data in digital formats despite the underlying phenomenon being continuous \cite{cheng2014managing}. The uncertainty in representing an object's movement is also impacted by the sampling frequency, or sample rate, at which location samples are collected \cite{pfoser1999capturing}. Furthermore, uncertainty within the data can accumulate from multiple sources and propagate into the model.

\subsubsection{Data uncertainty representation}

In a classification problem, the data uncertainty arises from the complexity of the data and the structure of the class boundary in the feature space. The uncertainty of the class variable, given a specific input instance, is quantified by the entropy of the true class distribution $\mathcal{H}[p(y|\boldsymbol{x})]$ where $y$ is the class label and $\boldsymbol{x}$ is the sample feature. The entropy reflects the randomness of the class distribution due to feature overlap among samples from different classes as shown in Fig.~\ref{fig:entropy_uq}.

For regression problems, data uncertainty arises from the inherent noise or variability in the data generation process. The observation noise, denoted as $\epsilon(\boldsymbol{x})$, is added to the true function $f(\boldsymbol{x})$ to obtain the target variable $y$. There are two types of noise: homeostatic noise, which assumes a constant observation noise across all inputs, and heteroscedastic noise, where the observation noise varies as a function of the input $\epsilon(\boldsymbol{x}) \sim p(\epsilon|\boldsymbol{x})$. The heteroscedastic noise model is useful when the noise level differs among samples.



\subsection{Model uncertainty}

\subsubsection{Source of model uncertainty}
Model uncertainty, also known as aleatoric uncertainty, encompasses the uncertainty in a model's predictions resulting from imperfections in the model training process. In the context of spatiotemporal data, model uncertainty can be attributed to three primary sources: uncertainty in model architecture, uncertainty in model parameters, and uncertainty due to dataset distribution mismatch. Uncertainty in model architecture arises from the lack of understanding regarding the most suitable model architecture for a given geospatial dataset. For instance, in deep learning models, uncertainty may exist regarding the optimal number of neural network layers and neurons in each layer, as an overly complex model can lead to overfitting. Uncertainty in model parameters stems from unknown optimal parameter values. This uncertainty can arise due to factors such as an improper training strategy, limited geospatial training instances, or convergence to a local optimum rather than the global minimum of the loss function. Therefore, the model's weight values may not accurately represent the true optimal solution. The last type of model uncertainty is caused by dataset distribution drift, where the distribution of test samples differs from that of the training dataset. This issue, known as out-of-distribution (OOD) data, is not uncommon in real-world spatiotemporal deployments, as the test cases often involve complex and diverse scenarios.

\subsubsection{Model uncertainty representation}

Representing model uncertainty in spatiotemporal data poses challenges due to its multiple sources. Different methods can be adopted to estimate and represent uncertainty associated with each type. For uncertainty stemming from model parameters, Bayesian neural networks (BNNs) are commonly used \cite{jospin2022hands}. BNNs assume a prior distribution over the model parameters and aim to infer the posterior distribution to reflect parameter uncertainty. Uncertainty arising from model architectures can be estimated using deep ensembles. This approach involves constructing an ensemble of neural network architectures, training each model separately, and generating predictions that form a distribution on the target variable. The variance of these predictions serves as an estimation of prediction uncertainty. Uncertainty resulting from dataset distribution mismatch can be assessed by considering the proximity of new test samples to the training samples. As the test sample deviates further from the distribution of training data, the model uncertainty increases.

In summary, model uncertainty in spatiotemporal data arises from misspecifications in model architectures, parameters, and dataset distributions. Depending on the specific application, one type of uncertainty may dominate over the other, necessitating tailored methods to address it. The recognition and appropriate handling of these uncertainties are crucial for robust analysis and interpretation of data and model outputs in various geospatial fields. 


\section{Methodolies}

In the following, the main intuitions and approaches of the three types are presented and their main advantages and disadvantages are discussed.

\subsection{Model uncertainty}
\subsubsection{Bayesian Neural Networks:}

The Bayesian neural network (BNN) incorporates a prior distribution $p(\boldsymbol{\theta})$ on the neural network parameters and learns the posterior distribution $p(\boldsymbol{\theta}|\mathbf{X}, \mathbf{Y})$ based on the training dataset in Eq.~\ref{eq:bayesian1}. However, analytically solving for this posterior distribution is intractable, and approximation methods are necessary for prediction in BNN.
\begin{equation}\label{eq:bayesian1}\footnotesize
    p(\boldsymbol{\theta}|\mathbf{X}, \mathbf{Y}) = \frac{p(\mathbf{Y}|\mathbf{X},\boldsymbol{\theta} )p(\boldsymbol{\theta})}{p(\mathbf{Y}|\mathbf{X})}
\end{equation}

One approach is to select an approximation $q_{\phi}(\boldsymbol{\theta})$ from a parameterized class of distributions $\mathcal{Q}$ to approximate the posterior. Popular optimization methods for selecting $q_{\phi}(\boldsymbol{\theta})$ include variational inference \cite{blei2017variational} and Laplace approximation \cite{friston2007variational}, both of which impose assumptions and restrictions on the form of the approximated posterior. However, these restrictions can lead to inaccuracies in predictions and uncertainty quantification. Another widely used approach is the MC dropout method \cite{gal2016dropout}, which is simple and easy to implement. It demonstrates that optimizing a neural network with a dropout layer is equivalent to approximating a BNN using variational inference on a parametric Bernoulli distribution. Uncertainty estimation is obtained by computing the variance across multiple stochastic forward predictions with different dropout masks. However, MC dropout tends to be less calibrated than other baseline uncertainty quantification methods in many benchmark datasets.

\subsubsection{Ensemble models}
Ensemble models are a powerful approach that involves combining multiple neural network models during the prediction process. By aggregating the predictions of individual models, an output distribution is formed. The variability in predictions among the ensemble models can serve as an indicator of model uncertainty, where a higher variance implies a greater degree of uncertainty. To capture this uncertainty stemming from various factors, several strategies for constructing ensembles can be employed, such as bootstrapping or combining different neural network architectures \cite{lakshminarayanan2017simple}.

\subsubsection{Sample-density aware neural networks}
The mentioned approaches do not effectively handle model uncertainty arising from low sample density, where samples lying far from the training set support may lead to overly confident predictions. To address this, various approaches have been developed to create sample density-aware neural networks capable of capturing model uncertainty in such scenarios. These approaches include Gaussian process modeling (e.g., kriging \cite{montero2015spatial}) for spatial data, deep Gaussian processes, and distance-aware neural networks. Distance-aware neural networks, inspired by Gaussian process models, aim to characterize uncertainty based on the density of sample features. They utilize the feature extraction capabilities of deep neural networks (DNNs) to learn a hidden representation $h(\boldsymbol{x})$ that reflects meaningful distances within the data manifold \cite{van2020uncertainty}.

\subsection{Data uncertainty}
Generally speaking, data uncertainty is represented by the entropy of the distribution $p(y|\boldsymbol{x}, \boldsymbol{\theta})$, where $\boldsymbol{\theta}$ is the neural network parameters. Existing approaches for learning this distribution can be classified into deep discriminative models and  generative models.
\subsubsection{Deep discriminative models}

Deep discriminative models can be further categorized as parametric or non-parametric based on the format of the distribution. To quantify data uncertainty, a discriminative model directly outputs a predictive distribution using a neural network. The distribution can be represented by a parametric model, which assumes a parameterized family of probability distributions (e.g., Gaussian or mixture Gaussian) whose parameters (e.g., mean and variance) are predicted by the neural network \cite{kendall2017uncertainties}. Alternatively, a non-parametric model does not make any assumptions about the underlying distributions and outputs a prediction interval \cite{pearce2018high}. Prediction intervals provide a lower and upper bound $[y_l, y_u]$, within which the ground truth $y$ is expected to fall with a prescribed confidence level of $1-\alpha$ (i.e., $p(y \in [y_l, y_u])> 1-\alpha$). However, a drawback is that standard optimization strategies may not be applicable.

\subsubsection{Deep generative models}

Deep generative models (DGMs) are capable of learning the intractable data distribution $p_{\text{data}}(\boldsymbol{x})$  in the high-dimensional feature space $\mathcal{X} \in \mathbb{R}^{n}$ from a large number of independent and identically distributed observed samples. To quantify DNN data uncertainty, the basic idea is to employ DGMs to learn the predictive distribution $p(y|\boldsymbol{x})$ using the conditional deep generative model (cDGM) \cite{sohn2015learning}. Uncertainty quantification models based on cDGMs aim to learn a conditional density over the prediction $y$, given the input feature $\boldsymbol{x}$. This involves learning a model $g_{\boldsymbol{\theta}}(\boldsymbol{z}, \cdot): \mathbb{X} \rightarrow \mathbb{Y}$, where the generative model $g(\boldsymbol{z},\boldsymbol{x})$ with $\boldsymbol{z}\sim p(\boldsymbol{z})$ approximates the true unknown distribution $p_{\text{true}}(y|\boldsymbol{x})$. The variability of the prediction distribution $p(y|\boldsymbol{x})$ is encoded in the latent variable $\boldsymbol{z}$ and the generative model. During inference, for any $\boldsymbol{x}\in \mathbb{X}$, we can generate $m$ samples with $y_i=g_{\boldsymbol{\theta}}(\boldsymbol{z}_i,\boldsymbol{x})$, where $\boldsymbol{z}_i\sim p(\boldsymbol{z})$. From these samples $\{y_i\}_{i=1}^m$, we can quantify the prediction uncertainty by measuring their variability.

\subsection{The combination of model and data uncertainty}
Many frameworks have been proposed to jointly consider both data and model uncertainty, aiming for more accurate uncertainty quantification. One straightforward approach is to select one method from each category and combine them within a single framework. For instance, Bayesian neural network (BNN) models like Monte-Carlo dropout or ensemble models can be merged with prediction distributions, where uncertainty can be obtained from the total variance of the prediction. However, such simple combinations often introduce significant computational complexity. Alternatively, other methods have developed evidential deep learning \cite{malinin2018predictive}, which integrates both data and model uncertainty within a single deterministic model using evidence theory. This approach offers computational efficiency, but it requires the design of new optimization strategies and may not be suitable for certain network architectures.


\section{Future Directions}

\subsection{Spatial Sample Density and Nonstationarity}
As discussed in the introduction, one major challenge of spatiotemporal data is spatiotemporal non-stationarity. The data distribution can vary from one region (time) to another region (time). Such phenomena can be characterized as out-of-distribution (OOD) data or spatiotemporal outliers.  Given a training data distribution $p(x)$, the OOD data are those samples that are either unlikely under the training data distribution or  outside the support of $p(x)$. Accurate detection of OOD samples is of paramount importance in spatiotemporal model generalizability. Because the space is continuous and the boundary of individual homogeneous sub-regions is implicit, the model needs to learn such spatial patterns in order to quantify uncertainty due to nonstationarity. Another relevant source of spaital uncertainty is due to sparse training samples in the geographical space. This is often due to limited sensor observations. Traditionally, Gaussian process has been widely used to quantify such spatial uncertainty in continuous space. However, for deep neural network models, new techniques are needed that consider sample density both in  the non-spatial feature space and in the geographic space. 

\subsection{Spatial Imaging and inverse problem}

The goal of the imaging process is to reconstruct an unknown image from measurements, which is an inverse problem commonly used in medical imaging (e.g., magnetic resonance imaging and X-ray computed tomography) and geophysical inversion (e.g., seismic inversion) \cite{fessler2010model}. However, this process is challenging due to the limited and noisy information used to determine the original image, leading to structured uncertainty and correlations between nearby pixels in the reconstructed image \cite{kendall2017uncertainties}. To overcome this issue, current research in uncertainty quantification of inverse problems employs conditional deep generative models, such as cVAE, cGAN, and conditional normalizing flow models \cite{dorta2018structured}. These methods utilize a low-dimensional latent space for image generation but may overlook unique data characteristics, such as structural constraints from domain physics in certain types of image data, such as remote sensing images, MRI images, or geological subsurface images \cite{sun2021physics}. The use of physics-informed models may improve uncertainty quantification in these cases. It's promising to incorporate the physics constraints for quantifying the uncertainty associated with the imaging process.

\subsection{UQ for physics-aware DNN models}
Many applications in the field of spatiotemporal data minging involve physical systems that can be described using principles such as partial differential equations (PDEs) governing diffusion processes. While deep learning has proven effective in extracting complex patterns from data, standard deep neural networks lack explicit incorporation of the underlying physics \cite{kashinath2021physics,he2022earth}. To address this challenge, physics-informed neural networks (PINNs) have been developed to integrate physical principles and domain knowledge into deep learning for consistent predictions \cite{krishnapriyan2021characterizing}. PINNs encode the governing equations as residual losses to guide the optimization of neural networks \cite{karniadakis2021physics}, enabling the incorporation of physical constraints during model training.

However, many physical systems exhibit non-deterministic or unknown underlying principles. Chaotic and stochastic behavior is inherent in these systems, as exemplified by weather forecasting and climate prediction, where small perturbations can lead to significant differences in the predicted state (known as the "butterfly effect") \cite{kashinath2021physics}. Even with deterministic initial conditions, the governing equations (e.g., PDEs) driving the system may be stochastic, described by stochastic differential equations. Uncertainty quantification is crucial for improving the reliability of predictions, especially under distribution shifts. The uncertainty in modeling physical systems can arise from multiple sources. First, the initial and boundary conditions may be non-deterministic, and the system itself may exhibit chaotic behavior \cite{wang2019deep}. Second, the underlying physical principles may not be fully known, or the parameters of the governing equations may be stochastic, leading to violations of conservation laws in imperfect systems \cite{yang2019adversarial}.

These cases highlight the presence of inherent data uncertainty in stochastic physical systems. Probabilistic models offer a natural way to model distributions and incorporate stochasticity and uncertainty into neural networks. However, quantifying uncertainty in PINNs poses specific challenges. It requires simultaneously considering the physical principles and their uncertainties. Incorporating physical constraints can help mitigate data and model uncertainty \cite{he2020semi}. Various sources of uncertainty may arise, including randomness in the physical system itself, measurement errors, and limited knowledge of the governing equations. One potential approach to uncertainty quantification in PINNs involves building probabilistic neural networks  that propagate uncertainty from multiple sources  based on deep generative models for structured outputs \cite{oberdiek2022uqgan}.

\section{Conclusion}
In this paper, we provides a brief overview of  UQ of deep learning for spatiotemporal data, including its unique challenges and existing methods. We particularly focus on the importance of uncertainty sources. We also identify several future research directions related to spatiotemporal data.

\bibliographystyle{ACM-Reference-Format}
\bibliography{ref}

\appendix
\end{document}